\documentclass[11pt,a4paper,final]{article}
\usepackage[hyperref]{acl2020}
\usepackage{times}
\usepackage{latexsym}

\setcitestyle{numbers}

\usepackage{microtype}
\usepackage{amsmath}
\usepackage{graphicx}
\usepackage{float}
\usepackage{multirow}
\usepackage{booktabs}
\usepackage{makecell}

\usepackage{natbib}

\usepackage{bigstrut}
\usepackage{balance}
\usepackage{bm}

\aclfinalcopy % Uncomment this line for the final submission
\title{BRAND CELEBRITY MATCHING MODEL BASED ON NATURAL LANGUAGE PROCESSING}

\author{Heming Yang$^{1,2}$ , Ke Yang $^3$, Erhan Zhang $^4$\\
$^1$College of Computer Science and Software Engineering, Shenzhen University, China\\
$^2$School of Media and Communication, Shenzhen University, China\\
$^3$College of Management, Shenzhen University, China\\
$^4$School of Software $\&$ Microelectronics, Peking University, China\\
$\left \{  yangheming2017, yangke2018\right \} @email.szu.edu.cn$\\
$zhangeh\text{-}ss@stu.pku.edu.cn$}

\begin{document}
\maketitle
\begin{abstract}
  Celebrity Endorsement is one of the most significant strategies in brand communication. Nowadays, more and more companies try to build a vivid characteristic for themselves. Therefore, their brand identity communications should accord with some characteristics as human and regulations. However, the previous works mostly stop by assumptions, instead of proposing a specific way to perform matching between brands and celebrities. In this paper, we propose a brand celebrity matching model (BCM) based on Natural Language Processing (NLP) techniques. Given a brand and a celebrity, we firstly obtain some descriptive documents of them from the Internet, then summarize these documents, and finally calculate a matching degree between the brand and the celebrity to determine whether they are matched. According to the experimental result, our proposed model outperforms the best baselines with a 0.362 F1 score and 6.3\% of accuracy, which indicates the effectiveness and application value of our model in the real-world scene. What's more, to our best knowledge, the proposed BCM model is the first work on using NLP to solve endorsement issue, so it can provide some novel research ideas and methodologies for the following works.\\
  Index Term: Celebrity Endorsement; Text Summarization; Matching Model
\end{abstract}
\vspace{1.5em}
\section{Introduction}

With the development of the Internet and the transformation of traditional business models, brand building  and communication have met new developing opportunities. The platforms of we media  provide rich resources and web traffic for increasing brand exposure. Aside from media, there was a spectacle of ‘Good writings make people copy them’ in ancient China, which means that something is popular and widely purchased by people. Today, this phenomenon appears and spreads rapidly in the form of news. At the beginning of 2021, the e-commerce live streaming presided over by Dilraba (a popular actress in China) achieved sales of more than 10 million in ten minutes. More than this, HUAWEI honor 9 endorsed by Hugo (a veteran actor in China) sold 1 million units within 28 days after the press conference. All these reveal the huge dividends brought by the rise of the fan economy and celebrity effect. More and more companies begin to pay attention to inviting celebrities with high traffic and high brand fit to endorse their products as well as create a brand image with personality characteristics. It makes us think about the relationship between brand image and celebrity characteristics, to make effective brand construction and improve product sales.

Early studies found that the endorsement effect will change with the product. For example, the difference of endorsement in the research was explained by using the concept of ‘fitness’ . The matching degree between celebrities and brands depends on the fitness between brands and celebrity images perceived by consumers. This study also shows that when celebrities have a high consistency image with products, they can improve the reliability of advertisers and celebrities  \cite{1} Itpointed out that brands and celebrities are nodes that are connected through advertising. When the fitness is high, it is helpful to establish and recall this connection; Instead, if there is no obvious specific connection between celebrities and brands, there will be a ‘vampire effect’, that is, consumers only remember celebrities, not the products or services they endorse, and celebrities will suck up the vitality of the products   \cite{2}. However, most of the previous studies stayed at the level of conjecture and demonstration, it is no in-depth exploration of how to match celebrities with brands. Only a small sample composing of 100 participants was used and limited to a specific brand. The study showed that successful brand promotion needs to keep the uniformity between brand personality and celebrity, but relevant mechanisms and methods on how to match the two have not been proposed \cite{3}. Based on previous studies and existing problems, we further explored the selection and strategy of celebrities by using Big Data and matching algorithms to better apply celebrity endorsement to brand communication.

Because of long texts and noise, we proposed a text summary module to obtain descriptive texts. Given these, we further proposed a matching module to calculate the matching degree between brands and celebrities, and then give a reasonable celebrity endorsement scheme. A substantial amount of research in celebrity endorsement used questionnaires to collect data related to artificial characteristics, such as brand consistency and celebrity consistency. For example,   7-point semantic difference scales were used to measure the consistency among consumers, celebrities, and brands \cite{4}. When studying the influence of celebrity consistency preference on voting intention, in addition to displaying the webpages of celebrities who support politicians on Facebook, the participants were also required to answer the consistency preference of celebrities and politicians with a 7-point semantic difference scale \cite{5}. Based on NLP technology, we creatively proposed an algorithm for analyzing descriptive texts. At the same time, considering the limitations of the questionnaire, we use the multi-dimensional experimental method to analyze experimental results. The BCM model in this paper exceeded all the baseline models and proved its effectiveness, which can provide some novel research ideas and methodologies for the following works.

First, we introduced the research background, theoretical and practical significance of matching celebrities and brands. Based on this topic, we investigated some related work including text summary algorithm, word vector model, and matching algorithm, and analyzes the characteristics of these algorithms. Then, we proposed the BCM algorithm based on NLP and collected experimental data to verify the effectiveness of the algorithm. In the end, we briefly summarized the work of this paper and looked forward to furthering research in the future.

The structure of the paper is as follows. The first Section mainly introduces the related work in the fields of text summarization algorithm, word vector model, and matching algorithm; the next Section is the structure and principle of the BCM mode; paragraph 3 carries out the experiment and introduces the experimental dataset, experimental setting, and experimental result; The last Section summarizes the main work of this paper and prospects the work in the future.

\section{RELATED WORKS}

\subsection{Text Summary}

Text summary algorithms can be roughly divided into two types: Extractive Summarization and Abstractive Summarization.
By calculating the importance of words or sentences in the text, the extracted text summarization algorithm selects a series of or the most important words and sentences to form the text summary. As early as 1958, Luhn \cite{6} proposed an extraction text summarization algorithm. Firstly, after removing the meaningless stop word, calculate the word frequency of each word as the importance of the word, and then select the first four sentences containing the most high-frequency words as the abstract of an academic paper or magazine article. BLEI \cite{7} et al. Proposed a late Dirichlet allocation (LDA) document topic generation model composed of a three-layer Bayesian probability model, calculated the topic probability distribution of the document and its sentences, and selected the most consistent sentences as the summary of the document. Cheng \cite{8} et al. Proposed a data-driven method, which uses a neural network and attention mechanism to extract abstracts based on sentence extraction or word extraction. The experiments of Cheng \cite{8} et al. Show that sentence-based extraction is better than word-based extraction. In recent years, the emergence of BERT \cite{9} has set off an upsurge of pre-training models in the field of natural language processing. Zhang \cite{10}and others also use rules to generate some pseudo labeled data based on the pre-training model and then carry out weakly supervised learning, to achieve the effect of state of art.
Although the model of Ruth et al. \cite{12} did not achieve good results better than the extraction text summarization algorithm, such seq2seq model aroused the research interest of researchers, and the model defeated many traditional text summarization algorithms, proving the feasibility of deep learning in the field of text summarization.
The following year, Chopra \cite{14} and others improved their algorithm, using recurrent neural networks (RNNs) with attention mechanism as decoder, and achieved better results. Li \cite{15} et al. Made a linguistic analysis of the artificial summary of CNN news and found that there are some fixed formats in the summary, which inspired them to design the module to integrate the potential structural information into the summary model. To achieve this goal, they added a latent structure model to the work of Chopra \cite{14} and others and used variational autoencoders (VAEs) \cite{20} as the generated framework. Since VAEs are not an end-to-end model, they propose a drug model (deep recurrent generative decoder) for text summary generation. Gehrmann \cite{17} et al. Proposed a bottom-up text summarization algorithm to decouple the text summarization into two steps: the first step is to mark the text sequence and select candidate words, and the second step is to use these candidate words to generate the text. Thus, the model can generate a more accurate summary. Liu \cite{18} and others proposed the bet sum model based on the pre-training language model BERT \cite{9}, which defeated the existing extraction text summary model and generative text summary model at that time and achieved the best results.
This paper also uses the generative text summary algorithm to generate text summaries for the descriptive documents corresponding to celebrity brands.
\subsection{Word Vector}
Different from images, natural language is discrete. How to represent natural language is a key problem. One-hot embedding is one of the most common word expressions. However, there are different degrees of similarity between words, such as "happy" and "happy", "sunshine" and "dark". It is obvious that one-hot embedding cannot express such properties in natural language. Mikolov \cite{19} et al. Proposed word2vec word vector model. Unsupervised training was conducted on the corpus-based on the skip-gram model and a continuous bag of word model, and then the distributed representation of words was obtained as the input of the downstream task model. Experiments show that after pre-training, the word2vec model learns the relevant information of words.
However, the word2vec word vector is static, that is, each word corresponds to only a one-word vector. Such a word vector cannot represent polysemy. Therefore, McCann \cite{20} et al. Proposed the CoVe word vector model, with the help of machine translation task, let the model complete the machine translation task, learn the contextualized word vector based on context, and generate the corresponding word vector according to different contexts. Peters \cite{21} et al. Proposed the Elmo word vector model composed of deep bidirectional long, short-term memory (LSTM), which can unsupervised learn the contextualized word vector of words through a more general language model than machine translation. With the help of a language model, large-scale training is possible.
In recent years, the proposal of transformer \cite{22} makes it possible for a large-scale pre-training language model. GPT \cite{23} and BERT \cite{9} are both transformer-based models. GPT carries out pre-training on large-scale data sets through a one-way language model, and finally obtains the representation of contextualized word vector. While BERT cleverly designed a masked language model (MLM) and next sense prediction (NSP), two pre-training tasks. After pre-training on a large corpus, the pre-training model was applied to downstream tasks, and the best performance of more than 10 tasks was obtained at that time. Due to the universality of the pre-training word vector model and its great help to the improvement of downstream tasks, more and more researchers are constantly improving the existing word vector model and putting forward new model architecture and new pre-training tasks \cite{24, 25, 26}. Baidu's Ernie \cite{27} recently broke glue (the benchmark for evaluating common language understanding) [ https://gluebenchmark.com/] The total score of emotional sentence classification (SST-2) and natural language inference task (WNLI) reached the first in the world.
Therefore, a good word vector model is of great significance for natural language processing tasks. This paper also uses the word vector model to encode the input text for further processing. When matching the generated abstracts, this paper also uses a word vector model to encode the abstracts and then carries out subsequent text matching.
\subsection{Text Matching}
Text matching algorithms are used in many fields, including text similarity calculation \cite{28}, question and answer matching \cite{29}, text implication \cite{30}, information retrieval \cite{31}, etc. When matching two texts, a natural idea is to calculate the cosine similarity between their representations. As the pioneering work of text matching, DSSM \cite{31} proposed a text-matching framework based on representation batted twin networks. Firstly, multi-layer perceptron (MLP) is used to extract the vector representation of a query and document respectively. Then, the cosine similarity of the vectors of the query and other documents are calculated respectively, which is sorted according to the similarity, and the document results corresponding to the query are provided to the user. Feng \cite{32} et al. Introduced CNN into the encoder so that the representation not only contains the global information of the text but also extracts its local information.

Different from the representation-based text matching framework of DSSM, Pang \cite{33} et al. Proposed a pyramid text matching model Matchpyramid, which simulates the process of image recognition and uses multi-layer CNN to carry out matching interaction hierarchically, to complete text matching. Chen \cite{34} et al. Proposed an ESIM model based on LSTM. To capture more detailed text information, Chen \cite{34} and others designed a fine sequential inference structure based on LSTM, realized local natural language inference through inter-sentence attention mechanism, and then further realized global inference. Because of its excellent performance in short text matching tasks, ESIM is often integrated into its text-matching model by contestants in many later competitions.

As a representation-based text matching model, DSSM has the advantage that it can pre-calculate the text, obtain the corresponding representation of the text, and realize the subsequent rapid text matching. It has great application significance in the field of information retrieval.

However, due to the lack of interaction with the target sentence, it cannot capture the features that are more conducive to matching. Interaction-based models such as Matchpyramid and ESIM can just make up for the shortcomings of the representation-based text matching model. However, too fine  and complex interaction modules will also make the inference speed unsatisfactory and cannot be well applied in practice. Nie \cite{35} et al. Proposed that DC-BERT considers the advantages of both and makes great use of the representation ability of the pre-training language model. DC-BERT uses BERT \cite{9} to encode documents and queries at the bottom and then uses a layer of transformer \cite{22} to interact with the encoded representation, to realize text.

Because this paper will first generate the summary of the document and then match it, that is, short text matching, this paper uses an interactive method for text matching.

\section{CELEBRITY AND BRAND MATCHING ALGORITHM}
In this chapter, we propose a celebrity brand endorsement matching algorithm combining text summarization and a word vector matching algorithm. In the following content, the theory and implementation details of each module in the algorithm will be introduced step by step.
\subsection{Problem Definition}
For a celebrity and a brand brain, use crawlers to crawl descriptive documents related to them on the Internet. We need to design a model $g(D^{(Celebrity)}, D^{(Brand)})$ to calculate the similarity between celebrities and brands according to a given document. These documents are shown in equations (1) and (2).

\begin{figure*}[htbp]
	\centering
	\includegraphics[width=0.95\textwidth]{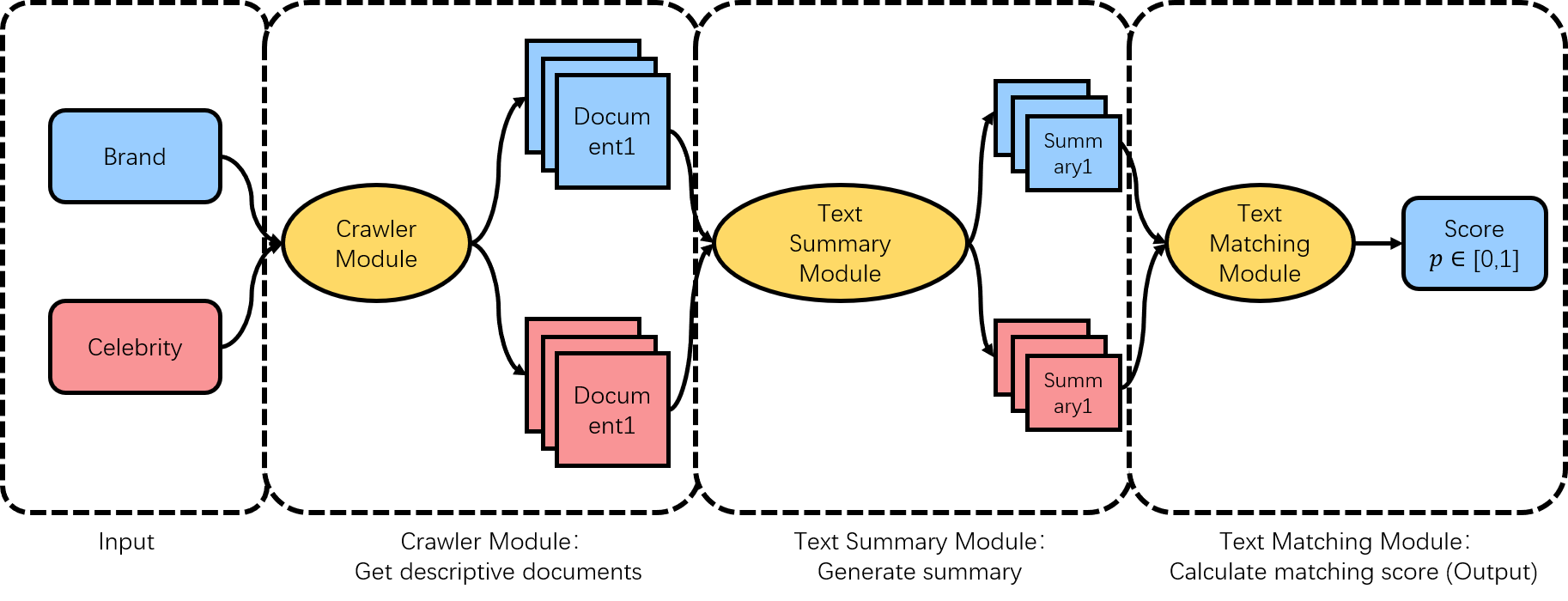}
	\caption{Model Structure of Celebrity Brand Matching Algorithm. The model proposed in this paper is mainly divided into three parts. The first part is the crawler module, which receives the input celebrity and brand names and crawls the relevant encyclopedia and news on the Internet; The second part is the text summary module, which generates the corresponding text summary for the obtained descriptive document; The third part is the matching module, which calculates the matching score between celebrities and brands according to the text summary of the descriptive documents corresponding to celebrities and brands, to judge whether a given celebrity and brand match.}
	\label{Fig. 1}
\end{figure*}

\begin{equation}
 \begin{aligned}
     &D^{(Celebrity)}=\\ &\left\{D_{1}^{(Celebrity)},D_{2}^{(Celebrity)},...,D_{k}^{(Celebrity)}\right\}
 \end{aligned}
 \end{equation}
\begin{equation}
 \begin{aligned}
\noindent D^{(Brand)}=\left\{D_{1}^{(Brand)},D_{2}^{(Brand)},...,D_{k}^{(Brand)}\right\}
\end{aligned}
\end{equation}

\subsection{Matching Model Between Celebrities and Brands}

Figure 1 is the pipeline architecture of the celebrity and brand matching algorithm proposed in this paper. The celebrity brand matching algorithm proposed in this paper is divided into the following parts. The first part is the crawler module. For the input celebrities and brands, the crawler is used to crawl the corresponding entry introduction and news text on the network as the descriptive documents of celebrities and brands. The second part is the text summary module. For each long document crawled, the text summary module is used to summarize the long document into one or several sentences to condense semantics and reduce noise for subsequent modules. The third part is the matching module. According to the summary of celebrities and brands generated in the text summary module, the similarity between them, that is, the so-called matching score is calculated to judge whether the celebrities and brands match.

\subsubsection{ Crawler Module}

For the description of celebrities and brands, we can crawl the descriptive entries in Baidu Encyclopedia. In addition, celebrities and brands often appear in entertainment news and other news, so we also crawl the news related to celebrities and brands. A series of descriptive documents related to celebrities and brands are obtained.
To simplify the description of the algorithm, in the subsequent modules, we assume that only one descriptive document is crawled for each celebrity and brand, that is, celebrity and brand brain correspond to each celebrity, as shown in equations (3) and (4).
\begin{equation}
  \begin{aligned}
   & D^{(Celebrity)}=\\ &\left\{w_{1}^{(Celebrity)},w_{2}^{(Celebrity)},...,w_{n}^{(Celebrity)}\right\}	
  \end{aligned}
  \end{equation}
 \begin{equation}
  \begin{aligned}
    D^{(Brand)}=\left\{w_{1}^{(Brand)},w_{2}^{(Brand)},...,w_{m}^{(Brand)}\right\}
 \end{aligned}
 \end{equation}

 Among them, $w_{i}$ represents the $i$-the word in the document, $n$ here represents the number of words in the document corresponding to celebrities, and m here represents the number of words in the document corresponding to brands.

After obtaining the corresponding descriptive document, we send it to the subsequent modules. Since the subsequent processing of celebrity documents and brand documents is the same, we will generally represent the document as 
$\bm{D=\left\{w_{1}, w_{2}, \cdots, w_{n}\right\}}$.
 \subsubsection{Text Summary Module}
 Abstract text is widely used in industry and academia because of its stability and controllability, but it only extracts the original sentences in the document, resulting in insufficient diversity. The generative text summary model can generate a variety of text summaries according to a given document. Therefore, this paper proposes a generative text summary model as the text summary module of the celebrity brand endorsement matching model. As the input of the text summarization module, this paper uses the word2ve word vector model to represent a text as several word vectors. Remember that the lookup Table of word2ve word vector model is, then for document $\bm{D=\left\{w_{1}, w_{2}, \cdots, w_{n}\right\}}$, we get its corresponding word vector matrix $\bm{E=\left\{e_{1},e_{2},\cdots,e_{n}\right\}}$ as the input of text summary module.

 This paper adopts the most common seq2seq model in Generative text summarization (also known as encoder-decoder structure) \cite{11}. As shown in Figure 2, the encoder part accepts an input sequence and encodes it. After encoding it into a hidden layer vector, the decoder generates the final output sequence one by one according to the information of this vector. To comprehensively consider each word in the input sequence when generating each word, attention is often used in some improved seq2seq models to calculate the weighted average of the previous words at each step for decoding.\\
 \begin{figure}[htbp]
	\centering
	\includegraphics{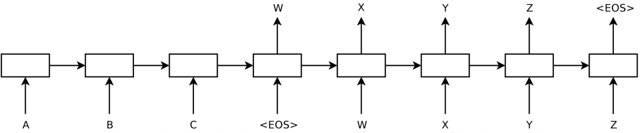}
	\caption{Seq2seq Model}
	\label{Fig.2}
\end{figure}
Since 2015, a lot of research work has used recurrent neural networks (RNN) to process sequence data. However, in recent years, more and more research work began to use transformer \cite{22} as the backbone model of seq2seq \cite{36, 10, 37}. This article also takes transformer as the basic component of our text summary module.
Figure 3 is the structural diagram of the transformer. It consists of an n-layer encoder and decoder. For each layer of the encoder, it includes a multi-head attention module and a feed-forward neural network. The multi-head attention module is composed of multiple scaled dot product attention. Figure 4 shows the calculation method of multiplicative attention.\\
\begin{figure}[htbp]
	\centering
	\includegraphics[scale=0.7]{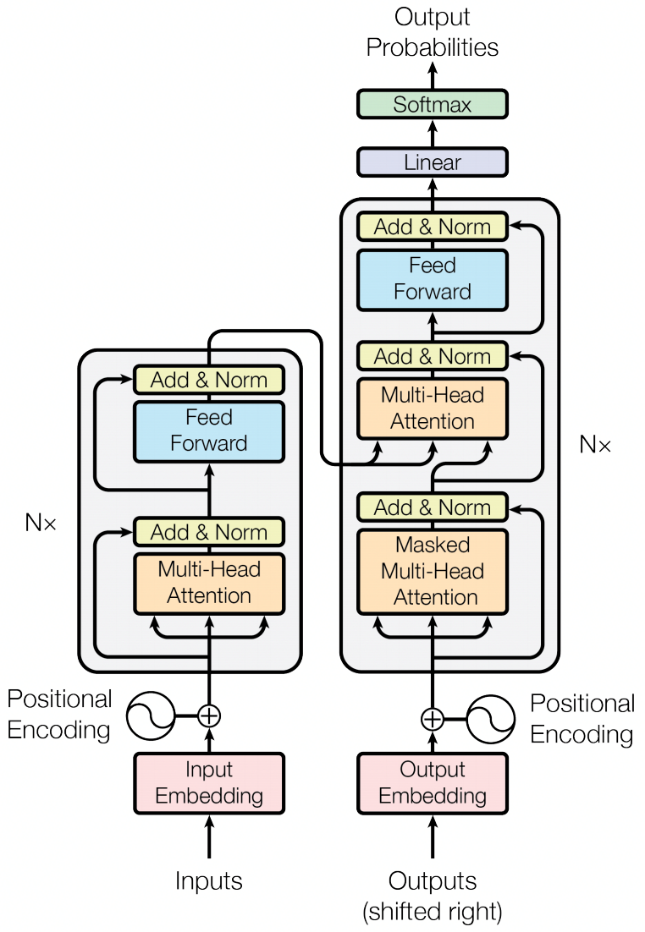}
	\caption{transformer \cite{22}.}
	\label{Fig.3}
\end{figure}
For input $q$, $k$, $v$, first perform linear transformation on them respectively, $\bm{W_{q}, W_{k}, W_{v}}$ is three trainable matrices, as shown in equation (5), equation (6) and equation (7):
\begin{align}
  Q=W_{q}q\\
  K=W_{k}k 	\\
  V=W_{V}v
  \end{align}

  Calculate their multiplicative attention, as shown in equation (8).
  \begin{equation}	
    \begin{aligned}
      Scaled Dot-&­Product Attention(Q,K,V)\\&=softmax(\frac{QK^{T}}{sqrt{d_{k}}})  
    \end{aligned}
  \end{equation}

  Where $d_{k}$ is the dimension of $K$.

  \begin{figure}[htbp]
    \centering
    \includegraphics{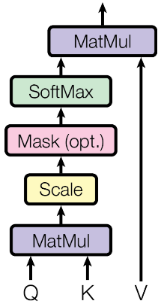}
    \caption{Seq2seq ModelScaled Dot-Product Attention \cite{22}.}
  \end{figure}

  Therefore, for the word vector matrix $\bm{E={e_{1},e_{2},\cdots,e_{n}}}$ of the document we input, we can also calculate its corresponding multiplicative attention, as shown in equation (9).

  \begin{equation}
    \begin{aligned}
      att_{E}=Scaled Dot­Product Attention(E,E,E)\\=softmax(\frac{EE^{T}}{sqrt{d_{E}}})E
    \end{aligned}	
  \end{equation}

 Among them, due to the input $q=k=v=E$, such an attention mechanism is also called the self-attention mechanism. Through this attention mechanism, the model can calculate the part that should be paid attention to according to the characteristics of the word vector, and then calculate a weighted average of the input to get a better representation. After the new representation is obtained, the residual connection is introduced and standardized, as shown in equation (10).
\begin{equation}	
  \hat{att}_{E}=Norm(E+att_{E}) 
\end{equation}

After the self-attention representation is obtained, it is input into the feedforward neural network, where $\bm{W}$ is the trainable matrix and $ \bm{b}$ is the trainable bias vector, as shown in equation (11).
\begin{equation}	
  Feed Forward(\hat{att}_{E})=W\hat{att}_{E}+b 
  \end{equation}

  Finally, after a residual connection and standardization, the output of the transformer encoder is obtained, and then the next decoding steps can be carried out, as shown in equation (12).
  \begin{equation}
    \begin{aligned}
      &Encoder(E)=\\&Norm(Feed Forward((\hat{att}_{E})+\hat{att}_{E})	
    \end{aligned}	
  \end{equation}

  The transformer decoder receives the output of the encoder and the corresponding decoder input. In this paper, we input the special character [CLS] into the decoder as the symbol for decoding and use $y_{0}=E_{[CLS]}$, $y_{i}$ as each output character, then the transformer decoder first calculates multiplicative self-attention, as shown in equation (13).
  \begin{equation}
    \begin{aligned}
      att_{y_{i}}&=Scaled Dot-Product Attention\\&(y_{i},(y_{i},(y_{i})=softmax(\frac{\sqrt{d_{y_{i}}}}{y_{i}y_{i}^{T}})	
    \end{aligned}	
  \end{equation}

  Then, the input of the encoder is introduced to calculate an interactive multiplicative attention, as shown in equation (14).

  \begin{equation}
    \begin{aligned}
        \text { att }_{y_{i+1}}=\text { Scaled Dot-Product Attention }\\ \left(E, y_{0: i}, y_{0: i}\right) 
        = \operatorname{softmax}\left(\frac{y_{0: i}^{7}}{\sqrt{d_{E}}}\right) y_{0: i}
    \end{aligned}	
  \end{equation}

  Finally, the final output is obtained through the feedforward neural network and standardization layer, and then the probability of each character is obtained through a SoftMax layer. In the training process, according to the output summary, we can compare with the real summary, calculate a cross-entropy error as the error of the whole neural network, and then back-propagation to update the parameters of each layer of the network. In the reasoning stage, we adopt the greedy strategy and take the character corresponding to the maximum output probability as the current output character for each step until the output terminator [SEP] or the maximum output sequence length is reached.

\subsubsection{Matching Module}
Based on the text summary module, we can get the summary $\bm{S_{i}=\left\{s_{1},s_{2},\cdots,s_{n}\right\}}$ of the document set corresponding to the celebrity or brand, where $\bm{S_{i}=\left\{i=1,2,\cdots,3\right\}}$ is the word in the summary. If there are multiple descriptive documents for each celebrity or brand entity, multiple document summaries will also be generated. For multiple abstracts, we splice these abstracts and then match them later. To speed up the calculation, we use a word2vec word vector to encode the words of the abstract, so as to obtain the corresponding word vector matrix $E_{s}$. For the word vector matrix of celebrity and brand abstract, we use $\boldsymbol{E}_{S}^{(\textbf {Celebrity })}$ and $\boldsymbol{E}_{S}^{(\textbf {Brand })}$  respectively.

Like the Matchpyramid model [33], we first calculate dot product for the word vector matrix of celebrity and brand abstracts, $\bigotimes$ representing dot product operation, as shown in equation (15).
\begin{equation}
  \begin{aligned}
    M=E_{S}^{(\text {Celebrity })} \otimes E_{S}^{(\text {Brand })}
  \end{aligned}	
\end{equation}

In this way, $m_{ij}$ represents the dot product similarity between the word $\boldsymbol{S}_{i}^{(\text {Celebrity })}$in the celebrity corresponding descriptive summary and the word $\boldsymbol{S}_{i}^{(\text {Brand })}$  in the brand corresponding descriptive summary.

In the field of image recognition, CNN is usually used as a feature extractor to extract different fine-grained information in the image \cite{38, 39, 40}. Similarly, we can also implement CNN on the similarity matrix $\mathbf{Z}^{(0)}=\mathbf{M}$ to extract local matching information between texts, as shown in equation (16).

\begin{equation}
  \begin{aligned}
    z_{i, j}^{(1, k)}=\sigma\left(\sum_{s=0}^{r_{k}-1} \sum_{t=0}^{r_{k}-1} w_{i, j}^{(1, k)} \cdot z_{i+s, j+t}^{(0)}+b^{(1, k)}\right).
  \end{aligned}	
\end{equation}

Among them, $z^{(i,k)}$ is the k-th feature map extracted by CNN on the first layer, and $\sigma $ represents the nonlinear activation function. In this paper, we use the ReLU \cite{41} function as the nonlinear activation function.

Then, each feature map will go through a pooling layer to further extract the local features in the feature map. Here, we select max pooling as our pooling layer to extract the maximum energy in the matching information, as shown in equation (17).

\begin{equation}
  \begin{aligned}
    z_{i, j}^{(2, k)}=\max _{0 \leq s \leq d_{k}} \max _{0 \leq t \leq d_{k}^{\prime}} z_{i \cdot d_{k}+s, j \cdot d_{k}^{\prime}+t}^{(1, k)}
  \end{aligned}	
\end{equation}

$z^{(2,k)}$ is the $ K-th$ feature map extracted from the pooling layer on the second layer.
Similarly, we can stack more CNN layers and pooling layers and finally get a highly generalized matching feature map. Then we flatten it and input it into the multi-layer perception (MLP) to get the final prediction result, as shown in equation (18).

\begin{equation}
  \begin{aligned}
    o=W_{2} \sigma\left(W_{1} z+b_{1}\right)+b_{2}
  \end{aligned}	
\end{equation}

Among them, p is the probability of celebrity and brand matching predicted by the model, $\bm{W_{1}W_{2}}$ is the trainable matrix, $\bm{b_{1}b_{2}}$ is the trainable bias vector, $\sigma$ represents the activation function, and ReLU is used as the activation function in this paper. Here is a 2-layer MLP model.

Because the celebrity and brand matching in this paper belongs to the two-classification task, and then through the sigmoid function, the final normalized celebrity and brand matching probability is obtained, as shown in equation (19).
\begin{equation}
  \begin{aligned}
    p=\operatorname{sigmoid}(o)=\frac{1}{1+e^{-o}}
  \end{aligned}	
\end{equation}

In the training process, the binary cross-entropy loss function is used as the loss function of the matching model, and  is the label of the -th training sample, as shown in equation (20).
\begin{equation}
  \begin{aligned}
    \text { loss }=-\sum_{i=1}^{N}\left[y_{i} \log \left(p_{i}\right)+\left(1-y_{i}\right) \log \left(1-p_{i}\right)\right]
  \end{aligned}	
\end{equation}

\section{EXPERIMENTAL DEMONSTRATIONS}

\subsection{Data Sets}

At present, there is no public data set related to celebrity brand endorsement, and there is no descriptive document data set specially designed for celebrities and brands, so all the data involved are crawled from the network. This paper selects 63 famous celebrities and 35 well-known brands for experiments. These celebrities and brands are more active, and the relevant descriptive documents will be relatively rich, so taking them as the research object can make our experiment more credible. We label all celebrity brands. If the celebrity has endorsed this brand within 5 years, the label is 1, otherwise, it is 0.

For the descriptive documents of celebrities and brands, we crawl their relevant descriptive documents and relevant news from Baidu Encyclopedia and today's headlines. Baidu Encyclopedia documents can describe the impression and information of a celebrity or brand, while Baidu news column documents can describe the impression and information of a celebrity or brand recently. Using these descriptive documents, we can not only comprehensively describe a celebrity or brand entity, but also consider its recent risk evaluation and image, to make the matching more reasonable and real-time. For each celebrity and brand entity, we crawl the corresponding Baidu Encyclopedia entry and 10 news documents.

For the crawled documents, we clean their data, including denoising, stop words, etc. The document statistics after preprocessing are shown in Table 1 below. The average number of words in celebrity-related documents is 1728, ranging from 28 to 16667; The average number of words in brand-related documents is 1572, ranging from 38 to 7841.

For each celebrity and brand entity, we randomly selected 1 Baidu Encyclopedia and 3 news documents for matching. To verify the effectiveness of the model, the star and brand texts are spliced, and the matching task can be regarded as a simple binary classification task. This paper selects a classical text classification model TextCNN to compare with this model.

TextCNN was first proposed by Kim \cite{42}. TextCNN obtains the feature representation of n-gram in sentences through one-dimensional convolution, which makes TextCNN have excellent

\noindent performance in text classification. TextCNN used in this paper is divided into four layers: embedding layer, convolution layer, pooling layer, and full connection layer. The input text vocabulary is converted into a 128-dimensional word vector and then processed by three types of 3 $\ast$ 128 convolution cores with sizes of 3, 4, and 5 respectively. The dropout probability in the network is set to 0.5. After 1200 rounds of training, the accuracy of the final training model is 59.6$\%$, F1 score: 0.485.
\begin{table*}[htbp]
	\centering
	\caption{Statistics of Crawled Documents}
	\resizebox{\textwidth}{15mm}{
	\begin{tabular}{|p{4.19em}|c|c|c|c|c|c|c|c|}
		\hline
		\multicolumn{1}{|r|}{\multirow{2}[2]{*}{}} & \multicolumn{4}{p{16.76em}|}{\makecell[c]{Baidu Wikipedia}} & \multicolumn{4}{p{16.76em}|}{\makecell[c]{Baidu News}} \bigstrut[b]\\
		\cline{2-9}    \multicolumn{1}{|r|}{} & \multicolumn{1}{p{4.19em}|}{\makecell[c]{Number }} & \multicolumn{1}{p{4.19em}|}{\makecell[c]{Average }} & \multicolumn{1}{p{4.19em}|}{Minimum} & \multicolumn{1}{p{4.19em}|}{Maximum } & \multicolumn{1}{p{4.19em}|}{Number } & \multicolumn{1}{p{4.19em}|}{Average } & \multicolumn{1}{p{4.19em}|}{Minimum } & \multicolumn{1}{p{4.19em}|}{Maximum } \bigstrut[t]\\
		\hline
		Celebrity & 63    & 4383  & 1061  & 16667 & 630   & 1413  & 28    & 9049 \bigstrut[b]\\
		\hline
		Brand & 35    & 3105  & 1007  & 7841  & 350   & 1419  & 38    & 6409 \bigstrut\\
		\hline
		Total & 98    & 3926  & 1007  & 16667 & 980   & 1415  & 28    & 9049 \bigstrut\\
		\hline
	\end{tabular}
	\label{tab:addlabel}
}
\end{table*}

\subsection{Evaluation Criteria}

\subsubsection{Classification and Evaluation Criteria}

The precision rate P (precision) and recall rate R were used in this experiment, F1 score and accuracy are used as evaluation criteria.
\begin{align*}
  \begin{gathered}
    \text { Precision }=\frac{T P}{T P+E P} \\
    \text { Recall }=\frac{T P}{T P+F N} \\
    F_{1}-\text { Score }=\frac{2 \times \text { Precision } \times \text { Recall }}{\text { Precision }+\text { Recall }}
    \end{gathered}
\end{align*}
TP:true positive\\
TN:true negative\\
FP:false positive\\
FN:false negative

\subsubsection{ROUGE}
The field of text summary usually adopts Rouge (recall-oriented understudy for getting evaluation) as the evaluation standard \cite{47}. It mainly counts the recall rate of the generated text summary, that is, the real summary and the generated summary are divided into several words, and then calculates how many of the generated summary is included in the real summary. The larger the size, the higher the quality of the generated text summary. The calculation formula is shown in equation (21)
\begin{equation}
  \begin{aligned}
   &ROUGE-N=\\ &\frac{\sum_{SE\mbox{{\small }}}\sum_{N-gramES}Count_{match}(N-gram)}{\sum_{SE\mbox{{\small }}}\sum_{N-gramES}Count_{match}(N-gram)}
  \end{aligned}	
\end{equation}
This paper mainly calculates the ROUGE-1 and ROUGE-2 indexes of the text summary module to judge whether the pre-trained text summary model works well. If the quality of the summary is poor, it will affect the matching effect of the subsequent matching model.

To evaluate the celebrity brand matching algorithm proposed in this paper, we learn from the experimental indicators of previous matching model research work, including precision P (precision), recall R (recall), F1 score (F1 score), and accuracy.

\subsection{Experimental Parameter Setting}

For the text summary module, this paper uses a 12-layer transformer as the backbone model, the maximum length of the position vector is 512, the hidden layer dimension of the encoder is 768, the maximum text length output by the decoder is 100, the number of multiple attention headers is 12, and the interlayer activation function is Gelu \cite{48}.

When training the text summary module, Adam optimizer \cite{49} is used to optimize the model parameters. The learning rate is set to $5\times$ $10^{-5}$. Each batch of training data contains 32 samples, and a total of 3 epochs are trained. To accelerate the convergence process, BERT \cite{9} is used to initialize the parameters of the transformer layer in the experiment.

For the matching module, this paper uses word2vec as the input word vector model. The matching network includes two layers of the convolutional neural network, two layers of pooling layer, and two layers of full connection layer. The convolution kernel size of the convolution neural network layer is $5 \ast 5 \ast 8$ and $3 \ast 3 \ast 10$ respectively. Except for the pooling layer, other layers take ReLU \cite{41} as the activation function.

Like the training text summary module, when training the matching module, the Adam optimizer is also applicable to optimize the model parameters. The learning rate is set to $1\ast 10^{-3}$. Each batch of training data contains 2 samples. After training 100 epochs, convergence is achieved. To evaluate the performance of the matching model, this paper divides the data set into 70\% and 30\% as training set and test set respectively.

\subsection{Experimental Results and Analysis}

\subsubsection{Text Summary Module}

After training on the training set in LCSTS and CLTS data sets, model evaluation is carried out on the pre-divided test set. The evaluation results are shown in Table 2. The value of ROUGE-1 is 0.5994 and the value of ROUGE-2 is 0.4259, which indicates that the text summary performs well.
\begin{table}[htbp]
  \centering
  \caption{Evaluation Indicators of Text Summary Module}
    \begin{tabular}{|c|c|}
    \hline
    \multicolumn{1}{|p{4.7em}|}{ROUGE-l} & \multicolumn{1}{|p{4.7em}|}{ROUGE-2} \\
    \hline
    0.5994 & 0.4259 \\
    \hline
    \end{tabular}
  \label{tab:addlabel}
\end{table}
To check the effect of text summary generation, Table 3 shows some text summary examples in the test set.

\begin{figure}[htbp]
  \centering
  \includegraphics[scale=0.85]{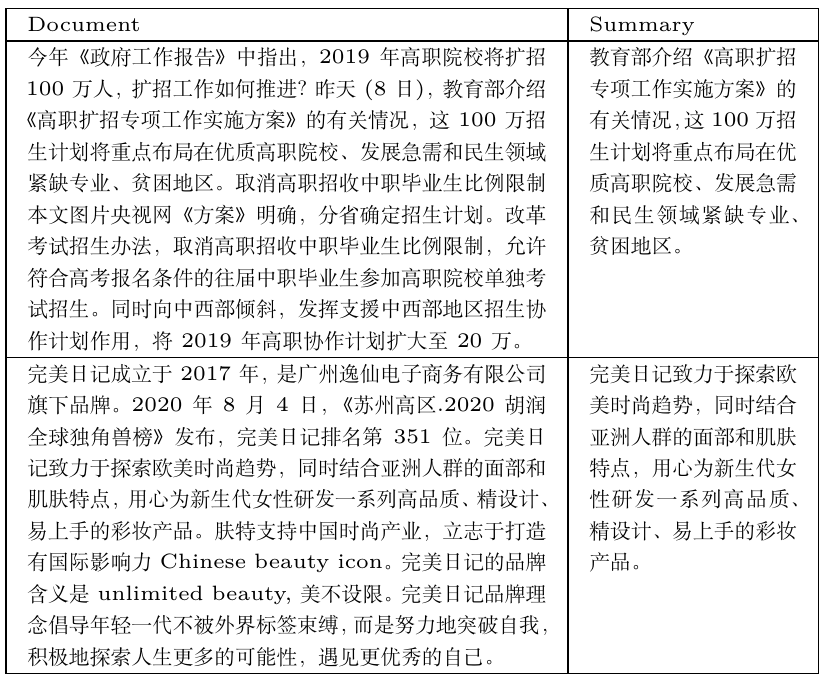}
  \caption{Sample Text Summary}
\end{figure}

% \begin{table*}[htb]
% 	\tiny
% 	\centering
% 	\caption{Sample Text Summary}
% 	\begin{tabular}{|p{25em}|p{10em}|}
% 	% \begin{tabular}{c|c}

% 		\hline
% 		% \multicolumn{1}{|c|}{Document} & \multicolumn{1}{c|}{Summary} \\
% 		Document & Summary \\
% 		\hline
% 		今年《政府工作报告》中指出，2019年高职院校将扩招100万人，扩招工作如何推进？昨天(8日)，教育部介绍《高职扩招专项工作实施方案》的有关情况，这100万招生计划将重点布局在优质高职院校、发展急需和民生领域紧缺专业、贫困地区。取消高职招收中职毕业生比例限制本文图片央视网《方案》明确，分省确定招生计划。改革考试招生办法，取消高职招收中职毕业生比例限制，允许符合高考报名条件的往届中职毕业生参加高职院校单独考试招生。同时向中西部倾斜，发挥支援中西部地区招生协作计划作用，将2019年高职协作计划扩大至20万。 & 教育部介绍《高职扩招专项工作实施方案》的有关情况，这100万招生计划将重点布局在优质高职院校、发展急需和民生领域紧缺专业、贫困地区。 \\
% 		\hline
% 		完美日记成立于2017年，是广州逸仙电子商务有限公司旗下品牌。2020年8月4日，《苏州高区.2020胡润全球独角兽榜》发布，完美日记排名第351位。完美日记致力于探索欧美时尚趋势，同时结合亚洲人群的面部和肌肤特点，用心为新生代女性研发一系列高品质、精设计、易上手的彩妆产品。肤特支持中国时尚产业，立志于打造有国际影响力Chinese beauty icon。完美日记的品牌含义是unlimited beauty,美不设限。完美日记品牌理念倡导年轻一代不被外界标签束缚，而是努力地突破自我，积极地探索人生更多的可能性，遇见更优秀的自己。 & 完美日记致力于探索欧美时尚趋势，同时结合亚洲人群的面部和肌肤特点，用心为新生代女性研发一系列高品质、精设计、易上手的彩妆产品。 \\
% 		\hline
% 	\end{tabular}%
% 	\label{tab:addlabel}%
% \end{table*}%

\subsubsection{Matching Module}

After the training of the text summary module, input the crawled documents to summarize them. In this experiment, each celebrity or brand entity has one Baidu Encyclopedia introduction and three corresponding headline news documents, so for each celebrity and brand entity, four abstracts of a certain length can be obtained.

In this experiment, the four abstracts are spliced into a long sentence for the training and testing of the matching model. Because the data set collected in this paper is unbalanced, and the proportion of positive examples is about 5.5\%, two benchmark models are designed to compare with the model proposed in this paper. One is random guess 0, which means that the consistent prediction is not matched during prediction; The other is random guess 1, which means that the consistent prediction is a match during prediction. The experimental results are shown in Table \ref{tab:result}.

\begin{table}[htbp]
	\centering
	\caption{Evaluation Index of Matching Module}
	\scalebox{0.7}{
	\begin{tabular}{|m{4.19em}|c|c|c|c|}
		\hline
		Model & \multicolumn{1}{m{4.19em}|}{{Precision}} & \multicolumn{1}{m{4.19em}|}{Recall} & \multicolumn{1}{p{4.19em}|}{F1 score} & \multicolumn{1}{p{4.19em}|}{Accuracy} \bigstrut\\
		\hline
		Random Guess 0 & 0     & 0     & 0     & 0.945 \bigstrut\\
		\hline
		Random Guess l & 0.055 & 1     & 0.105 & 0.055 \bigstrut\\
		\hline
		Text  & \multirow{2}[2]{*}{0.925} & \multirow{2}[2]{*}{0.511} & \multirow{2}[2]{*}{0.530} & \multirow{2}[2]{*}{0.926} \bigstrut[t]\\
		CNN   &       &       &       &  \bigstrut[b]\\
		\hline
		Our Model & \textbf{0.990} & \textbf{0.811} & \textbf{0.892} & \textbf{0.989} \bigstrut\\
		\hline
	\end{tabular}
	\label{tab:result}
}
\end{table}

It can be seen from the results in Table 4 that the model proposed in this paper greatly exceeds the other two benchmark models, and the precision rate reaches 99\%, which is much higher than other benchmark models; The recall rate reached 81.1\%, 30\% higher than TextCNN; F1 score is 0.892; The accuracy rate is 98.9\%, 4.4\% higher than random guess 0, which has the highest accuracy rate in the benchmark model, and 6.3\% higher than TextCNN, which is enough to illustrate the effectiveness of the model proposed in this paper.

In TextCNN, the way to splice and label two documents (Star+brand) to generate data requires that the algorithm itself can recognize the context structure between the two documents. TextCNN only carries out understanding and classification according to the topic of the text summary and cannot recognize the long-span context within the text. Therefore, the performance in performing different text mosaic classification tasks will be significantly worse than the model proposed in this paper.

As described in Section 3.3.1, the F1 score can comprehensively consider the precision rate and recall rate of the model. According to the results in Table 4 above, the F1 score of the model proposed in this paper is 0.892, which shows that the effect of the model proposed in this paper is very good. Its accuracy is 98.9\%, which also shows that its effect is very good.

\subsubsection{Analysis and Experiment}

For a celebrity or brand entity, the more descriptive documents corresponding, the more comprehensive and sufficient the relevant information. 
Various information can be considered in the matching, to improve the effectiveness of the model. Therefore, to explore the impact of the number of descriptive documents on the model effect, we evaluate the model effect when the number of documents is $\left \{1,2,3,4,5  \right \} $. Figure 5 shows the effect of the number of descriptive documents on the model effect.

\begin{figure}[htbp]
  \centering
  \includegraphics[scale=0.42]{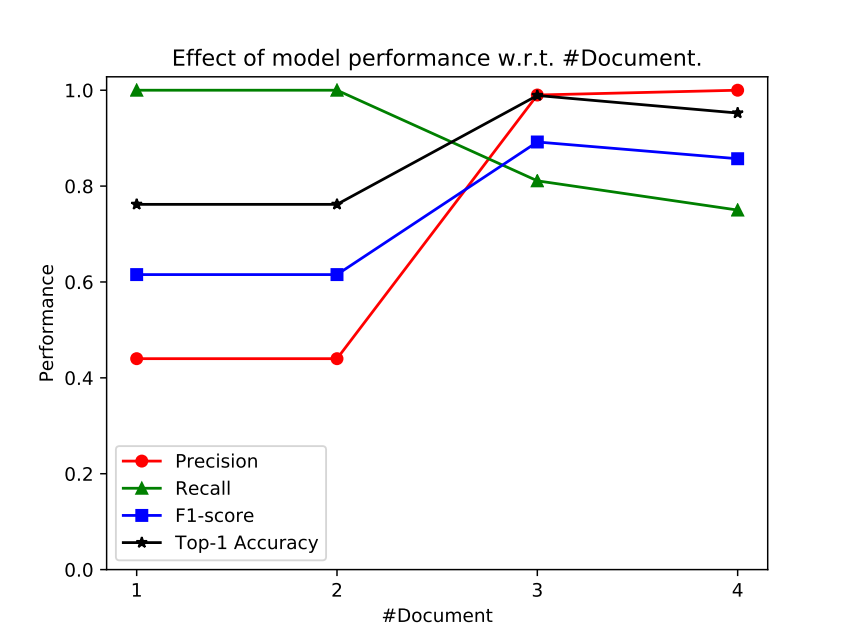}
  \caption{Effect of the number of descriptive documents on the model effect}
\end{figure}

The increase in the number of descriptive documents can indeed improve the effectiveness of the model. If only the summary of Baidu Encyclopedia is used as a match, the accuracy is only 23.44\% and the accuracy is only 82.13\%. However, if three related news articles are introduced and then matched, the accuracy rate is increased to 100\% and the accuracy rate is increased to 98.96\%. Although the recall rate dropped from 98.36\% to 81.15\%, the F1 score increased from 0.3785 to 0.8959, which shows that increasing the number of descriptive documents improves the overall effect of the model.

In addition, the model effect when the number of documents is 5 is consistent with that when the number of documents is 4, which indicates that increasing the number of documents after the number of descriptive documents reaches a certain value does not significantly help the matching.

\section{DISCUSSION AND CONCLUSION}
This study verified the better performance of the BCM model by comparing it with the benchmark models, and it can well identify whether a given celebrity and brand match.

Combined with Table 3, The BCM algorithm in this study will first generate a summary of a given descriptive text, which helps us better understand the intermediate variables of the model during text matching, observe how the model extracts effective information from long texts, and then complete the matching tasks. At the same time, if there is a problem in the text matching process, we can check the generated summary to find the source and cause of the problem. Compared with the previous matching methods, our algorithm flow effectively avoids the problem of fuzzy intermediate variables in the model and improves the interpretability of the model.

From the experimental analysis, we can see that the more descriptive documents and data about celebrities and brands, the more information that the BCM model can consider in the matching process, to effectively improve the accuracy of the algorithm. However, we need to note that the number of documents is not the more the better. When the number of documents reaches a certain value, adding documents will not significantly help the matching effect, and more computing resources will be required. Therefore, blindly increasing descriptive documents will just increase the calculation amount and operation time of the model.

We will consider using entertainment news to pre-train the model and improve the accuracy of the text summary model in the next. Besides, we collected a limited number of datasets due to the dispersion of celebrity endorsement information. In the future, we will continue to seek the services of entertainment companies, collect more celebrity brand endorsement data, and enhance the robustness and reliability of our algorithm.
\balance

\end{document}